\pdfoutput=1
\PassOptionsToPackage{breaklinks=true}{hyperref}
\PassOptionsToPackage{hyphens}{url}
\documentclass[11pt]{article}
\usepackage[final]{acl}
\usepackage{times}
\usepackage{latexsym}
\usepackage[T1]{fontenc}
\usepackage[utf8]{inputenc}
\usepackage{microtype}
\usepackage{enumitem}
\usepackage[font={itshape},rightmargin=0ex]{quoting}

\usepackage{booktabs}
\usepackage{graphicx}
\graphicspath{ {.} }

\usepackage{enumitem}
\setlist[description]{leftmargin=1em,itemsep=1pt}
\setlist[itemize]{leftmargin=1em,itemsep=0pt,topsep=0pt}

\title{Big AI is Accelerating the Metacrisis: What Can We Do?}
  
\author{Steven Bird \\
        Charles Darwin University\\
        Darwin, Australia}

\begin{document}

\maketitle
\begin{abstract}
    The world is in the grip of ecological, meaning, and language crises
    that are converging into a metacrisis.
    Big AI is accelerating them all.
    LLM engineering sits at the core.
    Despite the public good motives of language engineers
    and the promise of LLMs, this work is being leveraged to
    create unprecedented wealth and power for
    a handful of individuals and corporations
    while causing existential harm to life on earth.
    As a profession, we urgently need to come together to explore alternatives
    and to design a life-affirming future for our field of
    natural language processing that is centered on human flourishing on a living planet.
\end{abstract}

%%%%%%%%%%%%%%%%%%%%%%%%%%%%%%%%%%%%%%%%%%%%%%%%%%%%%%%%%%%%
\section{Introduction}
\label{sec:introduction}

So called ``Big AI'' \citep[p12]{Muldoon24} -- the corporations, the state capture,
the ``various kinds of automation sold as AI'' \citep[p162]{BenderHanna25} --
is escalating global crises which are reaching tipping point
and feeding each other in a \emph{metacrisis}.
Six of nine ``planetary boundaries'' have been breached \citep{Richardson23}.
There is a real prospect of ecosystem, economic, and geopolitical collapse \citep{Lenton23}.

\begin{figure}[t]
    \centering
    \includegraphics[width=\linewidth]{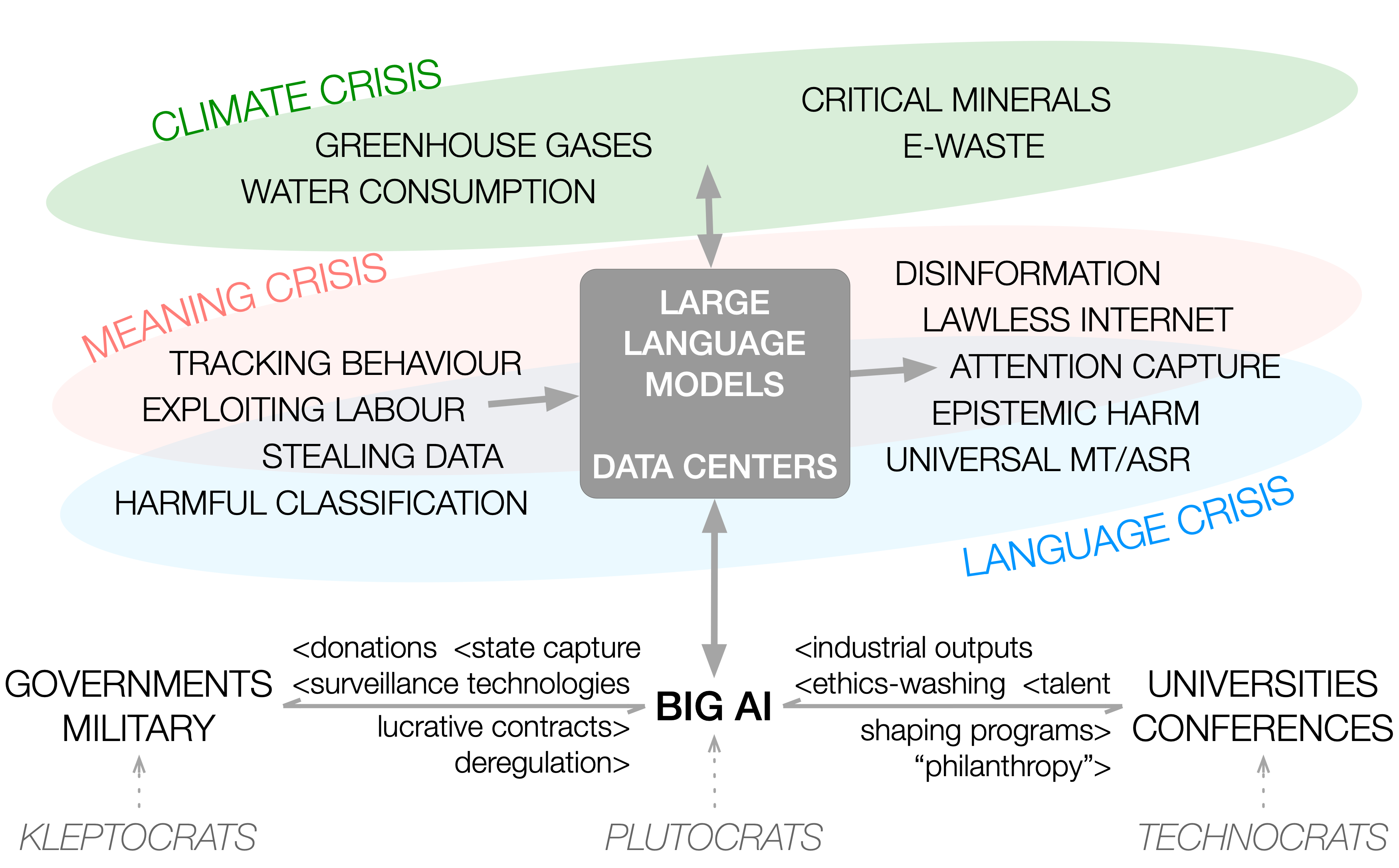}
    \vspace*{-3ex}
    \setlength{\belowcaptionskip}{-2.5ex}
    \caption{How Big AI is Accelerating the Metacrisis\label{fig:hpm}}
\end{figure}

Big AI fuels this system \citep[pp156ff]{BenderHanna25}.
Efficiency means scaling up, not saving resources \cite{Birhane22values,Fernandez25}.
Reductions in consumption attract new investment
in a rebound effect leading to greater consumption \citep{Alcott08,Weidinger22}.
Meanwhile, Big AI is ever ``reinventing itself in the public image as socially responsible, ...
influencing the events and decisions made by funded universities'' \citep{Abdalla21}.
The situation is set out in Figure~\ref{fig:hpm}.

The Association for Computational Linguistics (ACL)
might be the largest publisher of peer-reviewed LLM research.
Authors warrant that their work complies with the \emph{ACL Code of Ethics},
``understanding that the public good is the paramount consideration'' \citep{ACLEthics20}.
How are we to reconcile our professional obligation with the harms
caused by the technologies we are creating?

None of this is to blame individuals.
``We are all complicit.
We've allowed the `market' to define what we value so that the redefined
common good seems to depend on profligate lifestyles that enrich the sellers
while impoverishing the soul and the earth''
\citep[p307]{Kimmerer13}.
Brilliant scientists and engineers are being recruited into structures
that convert public goods such as Wikipedia into private goods like LLMs,
while extracting knowledge and behaviour through processes that have
been described as deception, exploitation, and theft \citep{Crawford21,Zuboff22}.
How, then, can we operate as language engineering professionals, dedicated to the public good as paramount?

In this policy paper
I compile evidence from a broad cross-section of literature to support the argument
that Big AI is accelerating the metacrisis~(\S\ref{sec:crises}).
I argue that much work under the heading of AI safety serves to perpetuate
an unregulated space where Big AI is able to continue without meaningful change (\S\ref{sec:assessment}).
Next, I make several suggestions for action at the level of the ACL (\S\ref{sec:what-can-we-do}).
Finally, I examine the logic of taking concerted action as a professional community (\S\ref{sec:community-standards}).\footnote{%
This paper is dedicated to William and Remy.}

%%%%%%%%%%%%%%%%%%%%%%%%%%%%%%%%%%%%%%%%%%%%%%%%%%%%%%%%%%%%
\vfil\clearpage
\section{Cascading Crises}
\label{sec:crises}

\subsection{LLMs are implicated in three crises}
\label{sec:three-crises}

When it comes to LLMs, three crises are particularly significant
thanks to the way language constitutes our common life:
our \emph{stewardship} of the planet;
the \emph{wellbeing} of our communities; and
the \emph{diversity} of our cultures.
I consider each in turn.

\paragraph{Ecological crisis:}

The world is experiencing a cascade of crises encompassing climate, pollution, and biodiversity,
linked to heatwaves, flash floods, drought and wildfires \citep{UNEP21,UNEP24}.
To this, data centres add excessive greenhouse gas emissions, e-waste, water usage,
and mineral extraction \citep{Crawford21,Kneese24,Kirkpatrick23,Schelenz22,UNEP24}.
Societal collapse is a plausible outcome due to
environmental perturbations that surpass the limits of adaptation,
the diminishing returns that come from increasing the complexity of social structures,
and the way risks propagate through complex systems \citep{Steel24,Kemp25}.

\paragraph{Meaning crisis:}
Big AI's business model is based on addiction \citep{Bhargava21}.
LLMs undermine critical thinking, 
creative work, knowledge diversity, and democracy \citep{Ressa22,Boninger25,Coeckelbergh25,Lee25,Katell25,Peterson25}.
LLMs perpetuate harmful classifications \cite{Bender21}.
LLMs generate fake news, education, and healthcare \citep{Zuboff22,Moore25}.
LLMs lack access to truth or social norms \citep{Hicks24,Resnik25}.

\paragraph{Language crisis:}

This is the collapse of the world's linguistic diversity
as minoritised speech communities experience persecution,
displacement, and genocide \citep{Crystal02}.
The problems are sociopolitical and are not solved by language technologies
\citep{Perley12,Bird20decolonising,Alvarado23,Helm24}.
Claims for the global utility of multilingual LLMs
neglect the realities that:
(a)~most of the world's population is multilingual, already using a few dozen contact languages for information access and economic participation;
(b)~90\% of the world's languages outside the most populous do not have standardised writing;
and (c)~there will never be enough data to create robust models for such languages
\citep{Hajek14,Kramer22,Bird24,Markl24,Moshagen24,Bird26sigul}.

\subsection{LLMs amplify crisis interactions}
\label{sec:metacrisis}

\paragraph{Ecological crisis $\leftrightarrow$ meaning crisis:}

The ecological crisis feeds the meaning crisis when
LLM content uses eco-anxiety to capture attention,
and doomscrolling numbs eco-anxiety \citep{Pearson24}.
In the reverse direction,
LLM content on social media narcotises dysfunction and apathy,
making it harder for communities to unite in the face of the ecological crisis \citep{Mateus20,Beato24}.

\paragraph{Meaning crisis $\leftrightarrow$ language crisis:}

The meaning crisis feeds the language crisis when
the avalanche of attention-grabbing LLM content from dominant languages crowds out local languages,
and when attention capture leads to non-participation in local lifeworlds \citep{Srinivasan17,Mateus20}.
In the reverse direction,
language loss undermines the place of elders,
disrupts knowledge transmission,
and harms wellbeing and cognition \citep{Perley12,Whalen16,Low22}.
The data free-for-all violates Indigenous (and human) sovereignty,
stoking both crises \cite{Walter19,Mahelona23,Bates25}.

\paragraph{Language crisis $\leftrightarrow$ ecological crisis:}

The language crisis feeds the ecological crisis by
undermining the capacity of indigenous communities
to take care of their ancestral lands which are rich in species diversity,
and when it accelerates the loss of medicinal knowledge as a resource for human wellbeing
\citep{Dieter05,Maffi05,Kimmerer13,Coyne22}.
In the reverse direction,
mining and climate disasters intensified by data centres
displace people from their lands (with their embedded stories),
while climate change and pandemics decimate linguistic communities,
and declining ecological diversity undermines cultures which depend on plant and animal species
\citep{Maffi05,Heugh17,Selvelli24,Ayuso25}.

\paragraph{The metacrisis:}

The world's crises are interconnected systems \citep{Wernli23},
drawing together in what has been called the polycrisis or the \emph{metacrisis} \citep{Morin99,Bhaskar16,Lawrence24},
``a complex system of interrelated, varied, and multi-layered crises'' \citep[p224]{Llena24}.
As shown, Big AI, its LLMs, and data centres are all implicated.
On top of this, Big AI accelerates itself through the phenomenon of AI hype \citep{LaGrandeur24,Markelius24,BenderHanna25}.
\vfil

\noindent
In short, \emph{Big AI is accelerating the metacrisis.}

%%%%%%%%%%%%%%%%%%%%%%%%%%%%%%%%%%%%%%%%%%%%%%%%%%%%%%%%%%%%
\vfil\clearpage
\section{More Business As Usual will not Work}
\label{sec:assessment}

\subsection{Big AI will not govern itself}

Big AI interest in ethics functions to minimise regulatory oversight \citep{Phan22,Zuboff22,Srivastava23,Yew25},
shaping governments and academia in
``a relational outcome of entangled dynamics between design decisions, norms, and power'' \citep[p724]{Shelby23}, cf.~Figure~\ref{fig:hpm}.
Big AI purports to ``solve'' the ethical problems of AI with more AI,
thanks to ``quantitative notions of fairness [that] funnel our thinking into narrow silos'' \citep[p158]{Mitchell21},
and leading to the devastating absence of the rule of law in the virtual world \citep[p4]{Ressa22}.

Politically-motivated ``philanthropy'' \citep{Bertrand20,Sherman25,Weissmann25}
and ethics washing \citep{Slee20,Seele22} function to maintain a deregulated space
in what is known as the ``mirage of algorithmic governance'' \citep{Slee20},
a ``fa\c{c}ade that justifies deregulation, self-regulation or market driven governance'' \citep[p267]{Bietti21}.
Big AI puts moneymaking before public safety \citep[p137]{Ressa22}.
``The idea that economic growth and the pursuit of profit should be tempered in the interests of individual well-being and less unequal societies is anathema to those in the vanguard of economic libertarianism'' \citep[p67]{Grayling25}.

There are many initiatives to regulate Big AI \citep{Wernli23,Bashir24},
but ``the social and environmental impacts of Gen-AI [are] complex and hinder targeted regulations'' \citep{Bashir24}.
``It will be far from straightforward to implement [AI ethics frameworks] in practice
to constrain the behaviour of those with disproportionate power to shape AI development and governance'' \citep[p576]{OhEigeartaigh20}.
However, this is no reason not to try.

\subsection{The scalability story is a myth}

Data centres cannot keep growing on a planet facing climate catastrophe 
(\citealt[pp156ff]{BenderHanna25}; \citealt{Hao25}).
AI safety is inherently not scalable \citep[p12]{Slee20},
and so we see a futile quest where guardrails are piled on top of monitoring systems on top of mitigations in a perpetual game of Whac-a-Mole.

Big AI's dirty secret is the annotation sweatshops located in the ``hidden outposts of AI'' \citep[p25]{Muldoon24}.
``The myth of AI as affordable and efficient depends on layers of exploitation, including the extraction of mass unpaid labor to fine-tune the AI
systems of the richest companies on earth'' \citep[p69]{Crawford21}.

\subsection{The benefits do not justify the harms}

In a consequentialist moment we might be tempted to dismiss the
manifold harms of Big AI (\S\ref{sec:crises}) considering the grandiose promises:
from ``eliminating poverty to establishing sustainable cities and communities
and providing quality education for all'' \citep{McKinsey24}.
However, since such benefits are imaginary, it is pointless to wait for them
and we should ask about the quality of the science itself.

Sequence models are far removed from \emph{natural} language
\citep{Bender20,Chrupala23,Bird24,Srivastava25}.
Much work is superficial and fashion-driven,
with SOTA-chasing and endless ``tables with numbers'' \citep{Church22,Kogkalidis25}.
Bias is generally understood ``as though it is a bug to be fixed rather than a feature of classification itself'' \citep[p130]{Crawford21}.
The actors are ``deafeningly male and white and technoheroic'' \citep[p9]{Dignazio23},
valorising technical novelty over all else \citep{Birhane22values}.
AI isn't working, or even helpful, for most people on the planet \citep[p143]{BenderHanna25}.

Only a minority of researchers can access SOTA tools,
and they must use exponentially more resources
for only linear performance gains \citep{Schwartz20}.
Review processes allow industry ``research'' to leverage the prestige of
conference publication into reputational benefits for private companies \citep{Young22}.
In spite of its manifold harms
LLM research receives scholarly recognition \citep{Abdalla23,Aitken24}.
Big AI is free to
``police its own use of artificial intelligence [leading to] the creation of a prominent conference on `Fairness, Accountability, and Transparency' [sponsored by]
Google, Facebook, and Microsoft'' \citep{Ochigame22}.
This move comes directly from the Big Tobacco playbook \citep{Abdalla21}.

These attributes of our field are not exactly hallmarks of ``good science''.
Thus, I find it difficult -- in a dispassionate assessment -- to conclude that
the benefits of Big AI to humanity and to science outweigh the significant risks
and manifest harms.

However, despite all of this, we can reclaim the commons.
The ACL community can do good LLM science and engineering.
We can recenter our work on the social, respecting humans as sovereign.

%%%%%%%%%%%%%%%%%%%%%%%%%%%%
\vfil\clearpage
\section{What Can We Do?}
\label{sec:what-can-we-do}

\paragraph{1. Public good as the paramount consideration.}
The ACL Code of Ethics applies to the conduct of members, not just our publications.
It is no excuse that ``someone's going to do it anyway''
(the myth of technological inevitability; \citealt{Mitchell20}), %p589
or that ``my contribution is but a small cog in a large machine''
(the problem of many hands; \citealt{FederCooper22}),
or that ``my work is connecting the world for good''
(the myth of technology solving social ills; \citealt{Srinivasan17}).
It remains a pressing open question for the ACL:
how do we reconcile the obligation of members to treat the public good as paramount
when our work is enabling Big AI?
Further, what do we make of the fact that Big AI research
involving human subjects is not subject to the kind of independent ethical oversight
that is required for university research?

\paragraph{2. Protect the ACL from corporate influence.}
The ACL's public good principle is not aligned with actors who ``understand AI as a commercial product
that should be kept as a closely guarded secret, and used to make profits for private companies''
\citep[p12]{Muldoon24}.
For example, Meta has been associated with a litany of evils from tax avoidance and addictive products
to election interference and genocide \cite{WikipediaFacebook,Goswami25,Schissler25,Taylor26}.
Why allow Meta to wash its image by sponsoring this conference?
Prima facie cases of apprehended bias, conflict of interest, and corporate capture
arise when Big AI employees participate in conference processes,
when they have greater opportunity to make publishable contributions
and, on account of access to SOTA infrastructure,
exert undue pressure on PhD research by students who want to maximise their employability
\citep{Schwartz20,Young22,Aitken24,Hao25}.

\paragraph{3. Shape the field of computational linguistics.}
The ACL has the liberty to
re-assert the scope of computational linguistics in its calls for papers,
centering the phenomenon of \emph{natural} human language
\citep{Bender20,Chrupala23,Bird24,Srivastava25}.
It can require deeper ethical engagement \citep{Liu23}.
It can encourage degrowth and small language models \citep{Vetter17,Meyers23,Wang25,Church26}.
It can use evaluation ``as a force to drive change'' \citep{Bommasani23}.
Such shifts may change what is considered prestigious research in NLP.

\paragraph{4. Establish protected spaces for critical NLP.}
A growing body of critical NLP research
reveals power dynamics and injustices
\citep[e.g.][]{Bender21,Markl22,Young22,Corbett23,Burrell24,Lopez24}.
I believe such work deserves a separate track and review process,
protected from reviewers who may guard the space for work that
maintains the status quo.

\paragraph{5. Establish spaces for NLP policy research.}
As global crises intensify,
the international community may unite behind new regulations
(cf.~Ozone, 1979; GFC, 2008; and COVID, 2019).
``Only when things get really bad ... is there a willingness to take decisive collective action.
The vulnerabilities are growing, and the necessary solutions are global...'' \citep{Bailey25}.
In anticipation, ACL could establish and promote a publication track for research on
policy that would inform regulatory activity.

\paragraph{6. Leadership with public statements \& policies.}

In view of the harms of LLMs, the ACL could provide informational statements
and policy positions \citep[cf.][]{ACLPolicy17,Goanta23,Kogkalidis25,Papagiannidis25,Schmitz25}.
This would serve ACL End~3 ``to represent computational linguistics to foundations and government agencies worldwide''
and End~4 ``to provide information on computational linguistics to the general public'' \citep{ACL00}.

\paragraph{7. Articulate a vision for life-sustaining research.}

What is our vision for language technology in the context of human flourishing
on a living planet?
New conference themes, workshops, and journal special issues are a start,
but we could promote new frameworks, e.g.:
proposals for resisting dehumanisation \citep{Bender24};
the Ethics of Care \citep{Cohn20,ElMasri25};
the 7 principles of data feminism \citep{Dignazio23};
the 10 Point Plan to Address the Information Crisis \citep[pp275ff]{Ressa22};
community-centric approaches \citep{BirdYibarbuk24,Cooper24,Markl24};
decolonising methods \citep{Smith12,Bird20decolonising,Mohamed20,Schwartz22,Lewis24};
and 5 steps for rewiring the machine \citep{Muldoon24}. %pp188ff
We could articulate values that go beyond ``public good'',
e.g.,~cognizance, beneficence, accountability, and non-maleficence \citep{Schwartz22},
or beneficence, justice, and inclusion \citep[p181]{Birhane22values},
or autonomy, creativity, ethics, slowness, and carefulness \citep[p14]{Bates25}.

%%%%%%%%%%%%%%%%%%%%%%%%%%%%
\vfil\clearpage
\section{Individual vs Community Standards}
\label{sec:community-standards}

As with computing in general, LLM narratives highlight
``the invention and perfection of the technology [which] downplays or disregards the contributions of the computer people'', aka ``computer boys'', who have ``constructed for themselves a unique occupational identity based on their control over the nascent technology'' \cite{Ensmenger10}. % pp4ff
No LLM harms would be possible without the tech bros
with their clubhouse mentality and a status hierarchy linked to their prowess with big machines
\cite[cf.][]{Margolis03,Ensmenger10,Dignazio23}.

How can members of our profession serve two masters,
one dedicated to the public good,
and one that serves ends that are antithetical to the public good?
%How realistic is it to expect them, that is, us, to now treat the public good as paramount?
Note that there is no recourse to the law as software engineering is not a recognised profession and
``ethical codes of conduct have little bite in determining legal duties of care'' \cite{Choi21}. % p24

Calling for individuals to comply with the ACL Code of Ethics is to expect them to
serve universal ethical principles regardless of peer expectations and local norms.
This is the highest level of moral development,
Kohlberg's Level 3 (cf.~Fig.~\ref{fig:kohlberg}).

\begin{figure}[b]
\vspace*{-2.05ex}
\hrule\vspace{1.5ex}\small
\begin{description}
\item[\textsc{Level 1: Preconventional}] \hfil \\[1pt]
  \emph{Rules and social expectations are external to the self.}
  \begin{itemize}
    \item avoid punishment and the superior power of authorities
    \item serve own needs in a world where others do the same
    \item (children<9, some adolescents, many criminal offenders)
  \end{itemize}
\item[\textsc{Level 2: Conventional}] \hfil \\[1pt]
  \emph{The self has internalised the rules and expectations of others, especially those of authorities.}
  \begin{itemize}
    \item living up to what is expected of people in one's role
    \item being seen as good person by self and by others
    \item (most adolescents and adults)
  \end{itemize}
  \item[\textsc{Level 3: Postconventional}] \hfil \\[1pt]
  \emph{The self is differentiated from the rules and expectations of others and adopts values in line with self-chosen principles.}
    \begin{itemize}
    \item commitment to social contract serving the welfare of all
    \item commitment to universal moral principles
    \item (a minority of adults)
    \end{itemize}
\end{description}
\vspace*{-3ex}
\caption{Kohlberg's Levels of Moral Development, with reasons for ``doing right'' (and typical membership),
adapted from \citep{Kohlberg76}}
\label{fig:kohlberg}
\end{figure}

It would be more pragmatic to assume conventional moral development (Level 2)
and to associate behaviour with membership of a prestigious group, ie.,~the ACL.
Then we can explore ways to develop, promote, and safeguard the expectations and norms of this group.
Thus, we anchor the idea that ``all of our flourishing is mutual'' \citep[p166]{Kimmerer13}.

\section{Conclusion}
\label{sec:conclusion}

The ``AI Gold Rush'' \citep{Greenstein20}
has become a ``silent nuclear holocaust in our information ecosystem'' \citep[p6]{Ressa22}.
Big AI is driving our world to the brink of collapse. % \citep{Lenton23,Richardson23}.

Meanwhile, language engineers persist with a scalability story that is failing humanity.
Our work consumes vast amounts of energy, critical minerals, and water,
and produces vast amounts of greenhouse gas and e-waste.
We participate in labour exploitation and data theft
in the name of SOTA-chasing, a dubious science.
We treat the finite natural world as an infinite storehouse and sewer,
while building language technologies as if the whole endeavour was apolitical and value-free.
For this, our professional body bestows recognition and prestige
which we parlay into grants and employment,
continuing the destructive cycle.

What can we do instead?
First, we can abandon the scalability ethic \citep{Hanna20},
recognising that true scalability comes through
humans supporting humans, perhaps assisted by language technology
\citep[cf. ``Technology's Law of Amplification'';][]{Toyama15}.
Anticipating that Big AI will be regulated or socialised \citep{Hanna20common,Varoufakis24},
and in line with ACL End~2 of promoting cooperation with related professional societies,
we can encourage research at the intersection of NLP and law \citep[cf.][]{Bestavros22}.

Second, our challenge is to
``rebuild our societies, starting from what's right in front of us: our areas of influence'' \citep[p2]{Ressa22}.
This includes our professional society with its social good intentions.

Finally, in enacting the social, we could view the ACL's Four Ends as social.
For example, End~1 of ``promoting research and development in computational linguistics''
is social because Language is relational.
This is not language for looking up or breaking down,
for spell checking or translating.

This is \emph{Language} as manifestation of human creativity,
\emph{Language} as ``spoken soul'' \cite{Rickford07}.
\emph{Language} for connecting people and country.
For sustaining our common life.

\section*{Acknowledgements}

This paper has benefitted from the thoughtful feedback of audiences of my CLARIN'24 and ALTA'24 keynotes,
plus many insightful reviews.
I thank Emily Bender for detailed comments
and Melissa Dwyer for advice on moral development.
Shortcomings remain my sole responsibility.

%%%%%%%%%%%%%%%%%%%%%%%%%%%%%%%%%%%%%%%%%%%%%%%%%%%%%%%%%%%%

\vfil\clearpage
\renewcommand*{\thefootnote}{\fnsymbol{footnote}}

\section*{Limitations}

\textbf{What does ``Big AI'' refer to?}
I have not identified individual corporations,
and I direct readers to lists compiled
by \citet{Abdalla21} and by \citet[pp11f]{Muldoon24}.

\textbf{Why not consider other crises?}
I have chosen only three crises,
a selection based on involvement by language and hence, LLMs (see the opening of \S\ref{sec:three-crises}).
I believe the approach could be applied to other crises.

\textbf{What about government-university links?}
Figure~\ref{fig:hpm} omits direct relationships between governments (including military) and universities.
The focus has been the central place occupied by Big AI.
The government-university axis is a locus for transformation:
both types of institution are constituted to serve society,
and both are able to regulate or sideline Big AI in the interests of society.

\textbf{Why target the ACL?}
I have identified the ACL as a locus of response for the NLP community,
and I have sought to engage the ACL leadership to learn about initiatives
that might already address some of the concerns raised here.
It might be argued that -- as a membership organisation --
the ACL is no more than the sum of its members, and limited to representing their interests.
I believe such a laissez-faire position is not justifiable in the face of the unfolding crises.
They merit a coordinated response at the level of our professional community.
By acting as a community, we and our elected representatives can
amplify our agency to bring about the future we would like to see.
This position is evident in the fact that the ACL has a code of ethics.
An easy first step would be to stop replaying hyperbole in our AI capabilities narratives \cite{EMNLP23,Zhang26}.
To be sure, it will be intergovernmental organisations (not examined here) that will finally regulate Big AI.
Nevertheless, we can regulate our publication channels
and foster expertise, norms, and policies that may eventually inform regulatory processes.

\textbf{Why bring politics to the ACL?}
An earlier version of this paper was criticised for ``courting controversy'' and being ``a political pamphlet''.
This neglects the fact that LLM engineering is already political and value-laden \citep[e.g.][]{Dotan20,Crawford21,Birhane22values,Coeckelbergh22,Markl22,Dignazio23,Markl25}.
Language is political, establishing and maintaining power structures \citep{Bourdieu91,Blodgett20}.
The conduct of science and technology is political \citep{Srinivasan17,Seguin23}.
The conduct of Big AI is political \citep{BenderHanna25,Hao25}.
The way LLM-generated content on social media platforms subverts democratic processes is political \citep{Jungherr23,Grayling25}.
Those who seek to banish overtly political content in the ACL context are being political:
\begin{quoting}
When people say ``that's too political'', or that they ``don't want to get political'', what they actually mean
is that they don't want to talk about politics that threatens a system they exist comfortably in...
And here's the greatest irony -- the people who don't realise how political everything is, those are the most political amongst us.
Because to be blessed with the privilege to move through the world without having to think about politics means that it's working
perfectly for you.\footnote[1]{\url{https://www.theblackproject.net/opinion-resources/like-it-or-not-everything-is-political}}
\end{quoting}

\section*{Ethical Considerations}

I disclose sponsorship by Google Research Australia
in the form of a small grant to Charles Darwin University in 2022,
on the understanding that I continue to discuss shared interests with Google staff.
I presented an earlier iteration of this paper
at a seminar at Google Research in San Francisco in May 2024, and
I gave input to the decision to select Aboriginal English as the target for
data collection work \citep{Hutchinson25}.
Google sponsorship has not affected the work reported here.

I have catalogued existential threats
which may be alarming for readers who do LLM research or work for Big AI companies.
Others have made related observations, e.g. on
the ``meaninglessness of SOTA-chasing'' \citep{Church22},
the ``problem of knowledge collapse'' \citep{Peterson25},
and the analogies with Big Tobacco \citep{Abdalla21}.
In view of what is at stake, some level of alarm is functional.

It would seem perverse to complain that those calling out harm should be censured for discussing unpleasant topics
and causing discomfort in readers.
Instead, let us ask:
\emph{what is the purpose of the ACL's Code of Ethics and its stressing of the paramount importance of public good
if members of the ACL community cannot call attention to practices that threaten existential harm to society and our planet?}

%%%%%%%%%%%%%%%%%%%%%%%%%%%%%%%%%%%%%%%%%%%%%%%%%%%%%%%%%%%%
\vfil

\end{document}